%% file: main.tex
\documentclass[letterpaper,10 pt, conference]{ieeeconf}  

\IEEEoverridecommandlockouts                              

\usepackage{epsfig} 
\usepackage{mathptmx} 
\usepackage{comment}
\usepackage{graphicx} 
\usepackage{subcaption}
\usepackage{caption}

\newcommand{\norm}[1]{\left\lVert#1\right\rVert}

\makeatletter
\let\NAT@parse\undefined
\makeatother

\usepackage{amsmath} 

\DeclareMathOperator*{\argmin}{arg\,min}

\usepackage{amssymb}  
\usepackage{pgfplots}
\usepackage{algorithm}
\usepackage{algpseudocode}
\usepackage{tabularx}
\usepackage{url}
\usepackage{amssymb}

\usepackage{cleveref}
\usepackage{cite}
\usepackage{etoolbox}
\makeatletter
\patchcmd{\@makecaption}
  {\scshape}
  {}
  {}
  {}
\makeatother

\usepackage[letterpaper,bindingoffset=0.2in,%
left=0.75in,right=0.75in,top=0.75in,bottom=0.75in,%
footskip=.25in]{geometry}

\usepackage{accents}

\usepackage{pgfplots}
\usepackage{lipsum}
\usepackage{xcolor}
\usepackage{pgfplots}

\title{
\LARGE \bf Environment-aware Interactive Movement Primitives for Object Reaching in Clutter 
\author{Sariah Mghames, Marc Hanheide
\thanks{The authors are with the University of Lincoln, UK, Centre for Autonomous Systems (L-CAS). This work is funded by Innovate UK \#51367, RobotHighways.}
}
}

\begin{document}
\maketitle
\thispagestyle{empty}
\pagestyle{empty}

\input{abstract}

\input{intro} \label{sec:intro}

\input{problem} \label{sec:problem}

\input{approach}  \label{sec:appr}

\input{experiments-vshort}  \label{sec:exp}

\input{conclusion} \label{sec:conc}

\section*{Acknowledgement}
The authors would like to thank Dr. Amir E. Ghalamzan for his insights to the problem and approach, Luca Castri for preparing the media material, 
and the EU H2020 Darko project for supporting the authors in concluding this work.

\bibliographystyle{IEEEtran}
\bibliography{IEEEabrv,references}

\end{document}

%% file: abstract.tex
\begin{abstract}
    The majority of motion planning strategies developed over the literature for reaching an object in clutter are applied to two dimensional (2-d) space where the state space of the environment is constrained in one direction. Fewer works have been investigated to reach a target in 3-d cluttered space, and when so, they have limited performance when applied to complex cases. In this work, we propose a constrained multi-objective optimization framework (OptI-ProMP) to approach the problem of reaching a target in a compact clutter with a case study on soft fruits grown in clusters, leveraging the local optimisation-based planner CHOMP. OptI-ProMP features costs related to both static, dynamic and pushable objects in the target neighborhood, and it relies on probabilistic primitives for problem initialisation. 
    We tested, in a simulated poly-tunnel, both ProMP-based planners from literature and the OptI-ProMP, on low (3-dofs) and high (7-dofs) dexterity robot body, respectively. Results show collision and pushing costs minimisation with 7-dofs robot kinematics, in addition to successful static obstacles avoidance and systematic drifting from the pushable objects center of mass.   
\end{abstract}

%% file: intro.tex
\section{Introduction}
Techniques for motion planning are presented over the literature to solve the problem of reaching a goal in an unstructured environment, hence solving a constrained problem by finding configurations in the free-space satisfying the kinematic limits of the autonomous system.
The majority of those techniques developed for applications of for e.g. navigation, pick-and-place and grasping, consider only the cost of vicinity to static obstacles, as the sampling-based techniques 
$Grasp-RRT$ in~\cite{vahrenkamp2012simultaneous} and the $RRT^*$-driven networks in~\cite{qureshi2019motion}. 
Other works have studied the possibility of formalising a more generic problem by accounting for the presence of both static and dynamic obstacles to find optimal trajectories, as the hierarchical reinforcement learning framework developed for interactive navigation in \cite{li2020hrl4in}, and for manipulation as in \cite{stilman2007manipulation} and \cite{wbejjani}. 

In this work, we approach the problem of reaching an occluded target in a highly cluttered environment by formalizing a generic multi-objective optimization framework that can be applied to 2-d (e.g. objects laying on a 2-d support) as well as 3-d state space (e.g. objects suspended in air), weighing the cost of the presence of both static and dynamic obstacles in the environment. The problem studied can be found in 3-d applications, e.g. agricultural robotics where manipulators are deployed in a poly-tunnel field for picking fruits grown in clusters, as well as in 2-d applications, e.g. the problem of reaching for a bottle of milk in a fridge or a box on the shelf of a warehouse.

\begin{figure}[t]
\centering
\includegraphics[scale=0.6]{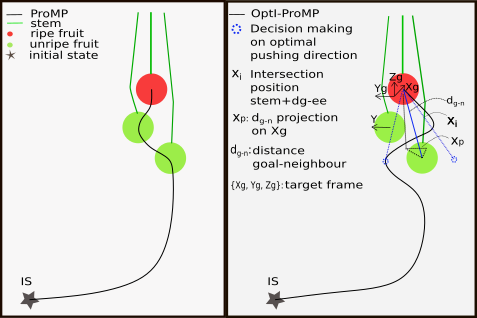}
\caption{OptI-ProMP: an optimization framework for reaching an object in clutter given environmental constraints, application to harvesting ripe fruits. Left figure represents the corresponding task space initialisation with a learnt probabilistic primitive conditioned at selected occluding elements (e.g. unripe fruits), at the desired target (ripe fruit) and at the initial pose. The right figure represents the desired optimized trajectory generated with a modified formulation of the Covariant Hamiltonian Optimisation for Motion Planning (CHOMP) problem setting. Variable $\mathbf{x}_i$ represents the intersection between stem samples consecutive to the end-effector pose with vector $\mathbf{d}_{g-ee}$.}\label{fig:intro}
\end{figure}

Latest advances in 2-d problems for object reaching in clutter propose deep reinforcement learning (DRL) \cite{wbejjani}-\cite{cheng2018reinforcement}, or optimization-based receding horizon techniques \cite{dogar2012planning}. 
In \cite{wbejjani}, the DRL approach relies on past experiences generated from a sampling-based algorithm ($RRT$ planner)
to learn a heuristic state-action value function which is then updated using the actual receding-horizon planner of the robot. Although their approach succeeds in finding an object in clutter, it lacks first learning from sub-optimal experiences and second is not readily applicable to 3-d problems where objects are closely inter-connected together and where sampling-based techniques may not converge to a solution. Similar limitations can be found in~\cite{dogar2012planning}.

On the other hand, 3-d problems such as reaching for a ripe fruit in a complex cluster of different coupled elements, is still an open challenge studied by many researchers, whether from a vision or planning perspective. The problem of cluttered suspension system (e.g cluster of fruits) faces 3 main challenges: (i) detection uncertainty caused by light intensity variation and vision system localisation accuracy, (ii) path obstructed by unripe fruits surrounding ripe ones and constrained by the manipulator kinematics, (iii) the growth of fruits near tables-top which increases the probability of collision with, while attempting to pick nearby. 
In \cite{xiong2019autonomous}, the authors have studied the problem of path planning to reach a ripe fruit in a cluster by advancing the end-effector in free space points next to unripe fruits. Although their approach presents high picking success rate, it suffers yet from the consideration of environment constraints (e.g. situation aforementioned in iii) which will most probably decrease the success rate due to the complex gripper design adopted (high dexterity with uni-directional scissor in a finger-like design), also it doesn't take into consideration the connection between objects to push, which is a major component of study when dealing with the identification of the pushing direction and with the challenge of obstruction diminution.\\    
More recently, the same challenge was addressed in \cite{mghamesinteractive} by proposing an interactive movement primitives algorithm to tackle the occlusion in clusters, taking into account an a-priori known stem orientation near the fruit lid. The approach presented in \cite{mghamesinteractive}, although relies on sub-optimal demonstrations from human expert for task learning and is computationally efficient, since primitive-based approaches can be computed offline and are adaptable online to new targets, yet it lacks the consideration of the obstacles in the environment. Also, the solution presented, although proved successful,
it relies on geometric analysis of the environment and hence is not an optimal end-to-end technique. \\
Other motion planning techniques adopted in more or less cluttered environments include but are not limited to, sampling-based techniques used for weed removal \cite{guzman2019weed} and for navigation~\cite{sharma2021}, deep reinforcement  learning~\cite{hoeller2021learning}, primitive-based approaches as~\cite{paraschos2013probabilistic}
for learning table tennis task and hockey game.\\
For 3-d cluttered problem settings, the authors in \cite{shyam2019improving} propose an optimization-based solution initialized with a probabilistic primitive for solving the problem of fast online trajectory generation for picking a ripe fruit (case of tomato grown in clusters). Although their approach shows efficient online time computation, it disregards the neighbourhood of the target. \\
In this work, we make inspiration from \cite{shyam2019improving} to build a probabilistic optimisation framework (OpI-ProMP, as shown in fig.~\ref{fig:intro}) for planning in cluttered environments with 3-d suspension systems by reasoning on both static and movable obstacles in the scene.

%% file: problem.tex
\section{Problem Formulation}

The task of reaching for an object in clutter, whether is a ripe fruit grown in a cluster of different elements (e.g. stems, foliage, etc)
or a common object on a shelf (e.g. cup in a cupboard), raises a question on the level of dexterity needed to achieve successfully the task. The higher the cluttering the more the need for dexterity. Another argument may arise when one considers the dexterity level. Shall the dexterity be embedded solely in the end-effector or the whole autonomous system. Dexterity can be thought to be embedded in either the control or the structure level, sometimes even in both. The more we concentrate the dexterity in one single system part, the larger the structure can be and hence the higher the difficulties to control that part intelligently so as to achieve successfully a task.
From here, one needs to think of a trade-off between body dexterity and end-effector dexterity. Nevertheless, increasing the dexterity in the body of an autonomous system leads to many considerations, among them the environment awareness and the computational complexities with the increased space of solutions to explore. 
\begin{figure}[t]
\centering
\includegraphics[trim=190 85 180 105, clip, scale=0.3]{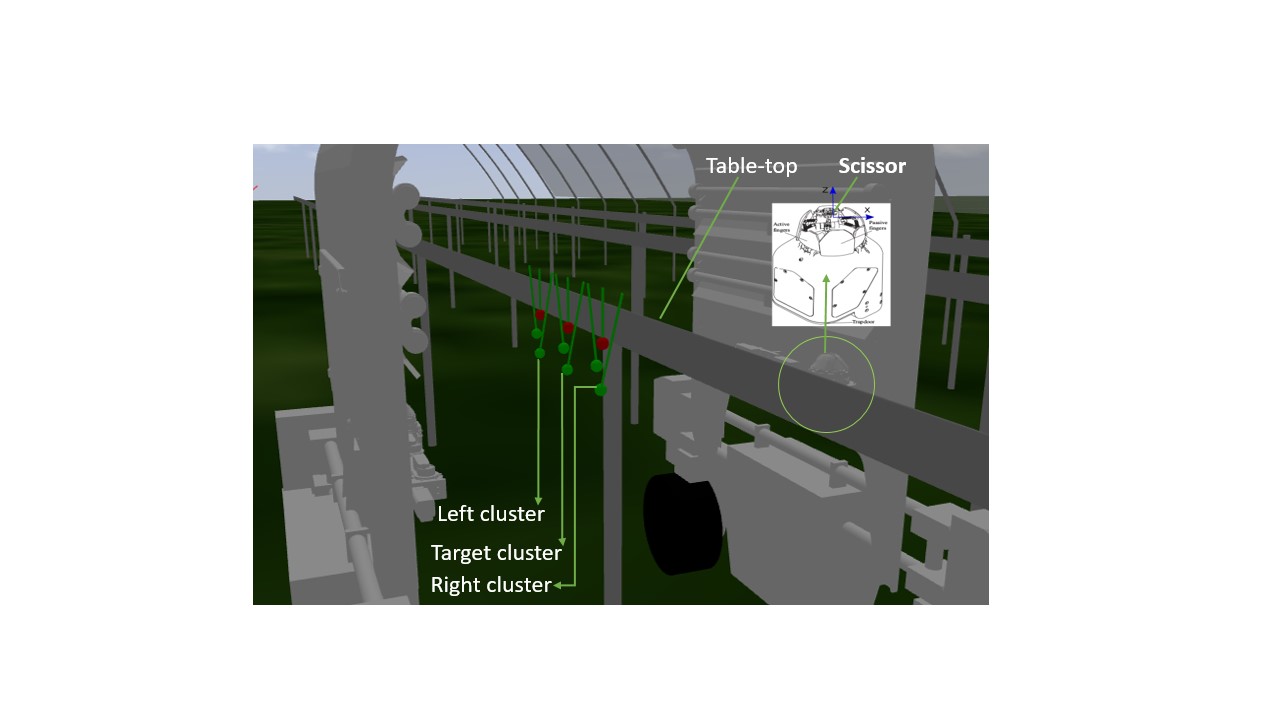}
\caption{Case scenario of environment-driven constraints with low-dexterity body and high-dexterity end-effector in a typical autonomous harvesting system deployed in real field: the Thorvald harvester.}\label{fig:prob}
\end{figure}

In our previous work\cite{mghamesinteractive}, we developed an interactive probabilistic primitive (I-ProMP) approach to deal with the problem of reaching a ripe fruit in a cluster. We tested our approach on Thorvald harvester system (\url{https://sagarobotics.com/}), shown also in fig. \ref{fig:prob}, mounted with two Scara arms, each with 3 degrees of freedom
(dofs), a finger-like gripper integrated  with  a  scissor  and  3  infra-red  sensors  to  localise  the  target inside. The latter work showed a working solution for few numbers of surrounding unripe fruits and with occluding elements (i.e. inclined stems). However, the work in \cite{mghamesinteractive}, didn't account for some case scenarios where environment constraints can impede the performance of harvesting with Scara arm given a highly-dexterous end-effector. For example, given the target cluster in fig.~\ref{fig:prob}, as it will be shown in sec.~\ref{sec:exp}, running the I-ProMP of \cite{mghamesinteractive} with a ripe fruit closer to the table-top, will lead eventually to a collision between the gripper and the table-top, and consequently to a task failure. This fact urges us to re-think a solution that can deal with scenarios of target adjacent to static obstacles. We propose to exploit the increase of structure dexterity in the robot body and decrease the dexterity level in the end-effector. Hence, we exploit in this paper the deployment of 7-dofs robot, Franka Emika, to reach for target in cluster, considering two kind of obstacles: (a) static obstacles as table-top, and (b) dynamic obstacles as the right and left hand clusters in fig.~\ref{fig:prob} surrounding the target cluster. We also leave the investigation into the minimum number of dofs needed to achieve high harvesting performance for a wide range of complex case scenarios, for future works.  

In the following, we present an approach to deal with increased dexterity in the body of an autonomous system tasked with reaching an object in a clutter.

%% file: approach.tex
\section{OptI-ProMP Framework}

To approach the problem of reaching in clutter while considering a broader range of complex scenarios, in this section we exploit the usage of higher dexterity level in the body of an autonomous system, as compared to the dexterity of its end-effector. For this, we select Franka Emika arm and we propose OptI-ProMP, an interactive optimisation framework that combines features of probabilistic primitives ($ProMP$) with the features of a well-known local motion planner, CHOMP. The selection of higher dofs manipulator induces the need to account for more types of obstacles due to the infinite number of solutions for a desired task space pose. Hence, we add a requirement constraint in OptI-ProMP that can penalise the cost of getting close to the surrounding dynamic objects (e.g. right and left hand clusters), decreasing therefore the possibility to damage natural elements or objects of use in our daily life.
\begin{figure}
\centering
    \begin{subfigure}[t]{0.43\columnwidth}
        \centering
        \fbox{\includegraphics[trim=680 240 433 176, clip, width=\textwidth]{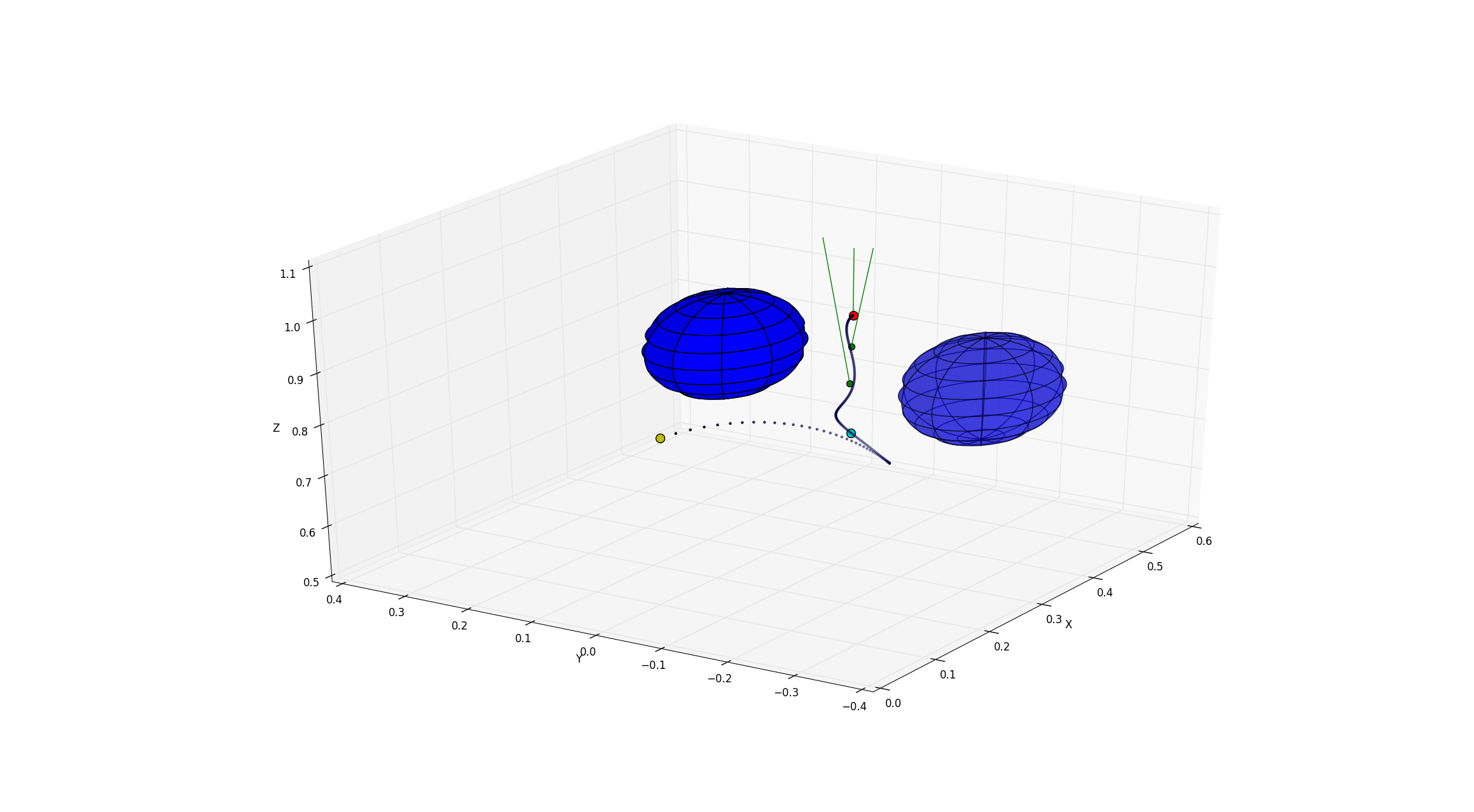}}
        \label{fig:promp_clus_obs}
     \end{subfigure}\hspace{10pt}
    \begin{subfigure}[t]{0.43\columnwidth}
        \centering
        \fbox{\includegraphics[trim=180 100 132 93, clip,width=\textwidth]{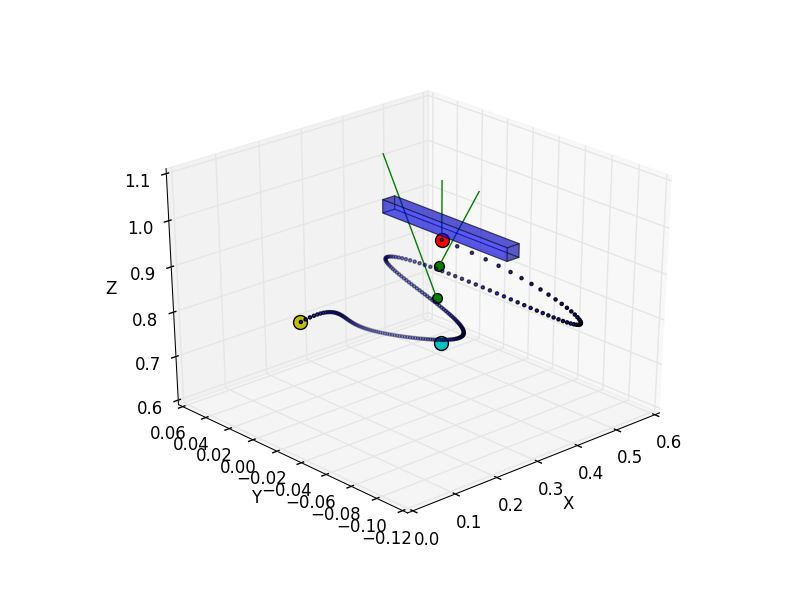}}
        \label{fig:promp_table}
    \end{subfigure}       
    \vspace{-10pt}
\caption{ProMP conditioning at initial robot state (IS, in yellow sphere), camera positioning state (cyan sphere), selected unripe pushable obstacles (case of fruits, in green spheres), and desired target pose (TS, in red sphere). Case scenarios of (right) a target cluster surrounded by other clusters (blue spheres of radius $10cm$) on its right and left hand side with $ProMP$ number of basis function $\psi = 5$ and width $h=4$, (left) a target cluster configuration (target fruit occluded by two pushable elements from below) given a target object in a close vicinity to a table-top, with $ProMP$ basis $\psi = 5$ and width $h = 5$. 
}\label{fig:cond_promp}
\end{figure}
We leverage the initialisation of OptI-ProMP on movement primitives which embed smoothness characteristic in their representation. Given this state initialisation, CHOMP doesn't need to consider smoothness penalty in its functional cost. Among the movement primitive variants, we follow a probabilistic approach ($ProMP$)~\cite{paraschos2013probabilistic} due to the uncertainty nature of the environments into consideration. The learnt $ProMP$ (as formulated in our previous work~\cite{mghamesinteractive}) is conditioned at the initial robot state, desired final state, state corresponding to camera pose beneath the target and outside the cluster, and the unripe objects, and used to initialise OptI-ProMP in each of the following environment settings: nearby clusters representing static obstacles (fig.~\ref{fig:cond_promp}-right), table-top representing a static obstacle right behind the target (fig.~\ref{fig:cond_promp}-left).

On the other hand, we propose that the objective functional optimising a given $ProMP$ penalises the probabilistic trajectory based on: (i) static obstacles interfering with the manipulator workspace, (ii) dynamic obstacles (considered static in the analysis) surrounding the target cluster, (iii) pushable/dynamic obstacles obstructing the path of an omni-directional gripper towards its target, (iv) interference of the manipulator main body (excluding final robot link to which the gripper is attached) with the dynamic obstacles obstructing the gripper path to target, (v) joint velocity measure. 
The velocity measure is introduced in the generic optimisation framework and we tested the necessity to apply it to agricultural case scenarios tackled in this work. The latter is driven by the need to ensure minimal generation of high velocities in the region around the target cluster, with robot configurations close to singularity. 

In the following we denote by $\xi$ the trajectory vector at the joint level, $F^{obs}_{s}$ the cost functional of the static obstacles, $F^{obs}_{d}$ the cost functional of the dynamic obstacles that we want to avoid, $F^{obs}_{p}$ the cost functional of the dynamic obstacles we want to push, and $F_{v}$ is the robot joint velocity cost functional.

\subsubsection{Obstacle Cost Functional}

\begin{equation} \small
    F^{obs}_{s,d}(\xi(t)) = \int_0^T \underset{u_s \in B_s, u_d \in B_d}{\max} \, c_{s,d}(x(\xi(t),u_{s,d})) \left\| \frac{d}{dt} x(\xi(t),u) \right\| dt
\end{equation}

where $c_{s}()$ is the cost with respect to the static objects (e.g table-top) and $c_{d}()$ is the cost with respect to the dynamic objects (e.g neighbour clusters). $\mathcal{B}_s$ is the set of body points of the n dofs robot and the end-effector link, whereas $\mathcal{B}_d$ is the set of body points of the (n-1) dofs robot. Adding $c_{d}()$ constrains the dynamics introduced in each cluster to the end-effector only, 
reducing therefore the variations in the initially detected cluster configuration. $x(\xi, u)$ is the workspace location of $u_{s,d}$ where $u_{s,d} \in \mathcal{B}_{s, d}$.

\begin{itemize}
\item \textit{Obstacle Cost Formulation: \textbf{$c_{s,d}()$}}
\newline
Hereafter, we define the obstacle cost function which penalises the robot for being near obstacles. As in~\cite{zucker2013chomp}, we define the cost function $c_{s,d}: \mathbb{R}^w \rightarrow \mathbb{R}$ in the robot's workspace as 
\newline 
\begin{equation}\small
    c_{s,d}(x) = \begin{cases}
    -D(x) + \frac{1}{2} \epsilon_s, \quad if \, D(x) < 0 \\
    \frac{1}{2\epsilon_s} (D(x)-\epsilon_s)^2, \quad if \, 0 < D(x) \leq \epsilon_s \\
    0 \quad  otherwise
    \end{cases}
\end{equation}
where $\epsilon_s$ is a safety parameter set to 1cm, and $D(x)$ is a signed distance field which stores the distance from a point $x \in \mathbb{R}^3$ to the boundary of the nearest obstacle. $D$ has negative values inside obstacles, positive outside, and zero at the boundary.
\end{itemize}

\subsubsection{Pushing actions Cost Functional}
\begin{equation}
 F_{p}^{obs}(\xi(t)) = \int_{t_1}^T \norm{fk(\xi_t)- X_g}^2_{\mathcal{F}_g} dt
\end{equation}
where $fk() \in \mathbb{R}^6$ is the forward kinematics vector, $X_g$ is the goal pose, and  $F_{p}^{obs}$ is the L2 norm of the projection of the vector distance between goal and end-effector (ee), on ${F}_g$, where $\mathcal{F}_g = \{x_g, y_g, z_g\}$ is a fixed reference frame attached to a target fruit. $F_{p}^{obs}$ is minimised form $t_1$ to $T$ where $t_1$ is the time corresponding to the conditioned camera positioning point. 
\subsubsection{Velocity Cost Functional}
\begin{equation}
F_{v}(\xi(t))= \int_0^T \dot{\xi_t}^T \, \xi_t \, dt
\end{equation}

where $\dot{\xi_t}$ is the joint velocity vector. Given the cost functional of joint velocity, obstacles collision, and pushing actions, respectively, we can now formulate the generic version of a constrained optimisation problem that minimises the total cost functional ($F$) given the initialised conditioned primitive. Hence, the problem becomes,

\begin{subequations}\footnotesize
\begin{alignat}{2}
\xi^*&= \argmin_{\xi(t)} \alpha_1 F_{s}^{obs}(\xi(t)) + \alpha_2 F_{d}^{obs}(\xi(t)) + \alpha_3 F_{p}^{obs} (\xi(t)) + \alpha_4 F_{v} (\xi(t))   \label{eq:optProb1}\\
\textrm{s.t.} & \quad \mid fk(\xi_t)- X_g \mid_{\mathcal{F}_0^g} \geq R_{gripper}, \quad t_1 \leq t \leq T 
\label{eq:constraint11}\\
& \quad \{p_1 \in \norm{fk(\xi_t)- X_g}\} \cap \{p_2 \in \mathcal{C}_C\} = \emptyset \label{eq:constraint12}\\
& \quad \delta [fk(\xi_t)]_x \quad \& \quad \delta [fk(\xi_t)]_z = 0\pm \label{eq:constraint13}\\
& \quad [q, \dot{q}]_{min} \leq \xi_t \leq [q, \dot{q}]_{max} \label{eq:constraint14}
\end{alignat}
\end{subequations}

Constraint~\ref{eq:constraint11} ensures the drift of the ee from the object center by a minimum distance of gripper radius. Constraint~\ref{eq:constraint12} ensures the drift is in the direction that has minimum intersection points ($x_i$ in fig.~\ref{fig:intro}) with the consecutive stem connection, constraint~\ref{eq:constraint13} ensures that the updated ee pose has little variation along the $x_g$ and $z_g$, while constraint~\ref{eq:constraint14} ensures the joints limits are satisfied. For the scenario of fruit clusters, $\mathcal{C}_c$ denotes the closest \textit{connection} and connections constitute the set of stems connecting each fruit to the environment.
The optimisation framework formulated in Eq. \eqref{eq:optProb1} can be generalised to 2-d pushing scenarios where \textit{connections} or couplings are in most cases not present.

We solve a nonlinear constrained optimisation problem locally, using the gradient descent technique, as follows
\begin{equation}
\xi^* = \xi - \gamma \bar{\nabla} F(\xi(t))
\end{equation}
The functional gradient for each of elemental cost can be calculated from the following derivation of the generalised formula, but the reader can refer to \cite{zucker2013chomp} 
for a detailed formulation.
\begin{equation}
\bar{\nabla} F(\xi) = \frac{\partial v}{\partial \xi}-\frac{d}{dt}\frac{\partial v}{\partial \dot{\xi}}
\end{equation}

where $F(\xi)$ can be written in the form $ \int f(\xi(t), \dot{\xi}(t))dt$.\\

%% file: experiments-vshort.tex
\section{Experiments: Simulation Environment}
In this section we present four real-world case scenarios on which we tested ProMP, I-ProMP, and OptI-ProMP planning frameworks. In OptI-ProMP, we choose $\{\alpha_1, \alpha_2, \alpha_3\} = 1$, while $\alpha_4 = 0$. \\
\newline
\paragraph*{\textbf{\textit{A. Scenario-I}}: \textbf{ProMP for picking fruits in cluster with Scara arm}}

\begin{figure}
\centering
     \begin{subfigure}[t]{0.4\columnwidth}
     \centering
         \includegraphics[trim=220 320 480 53, clip, width=\textwidth]{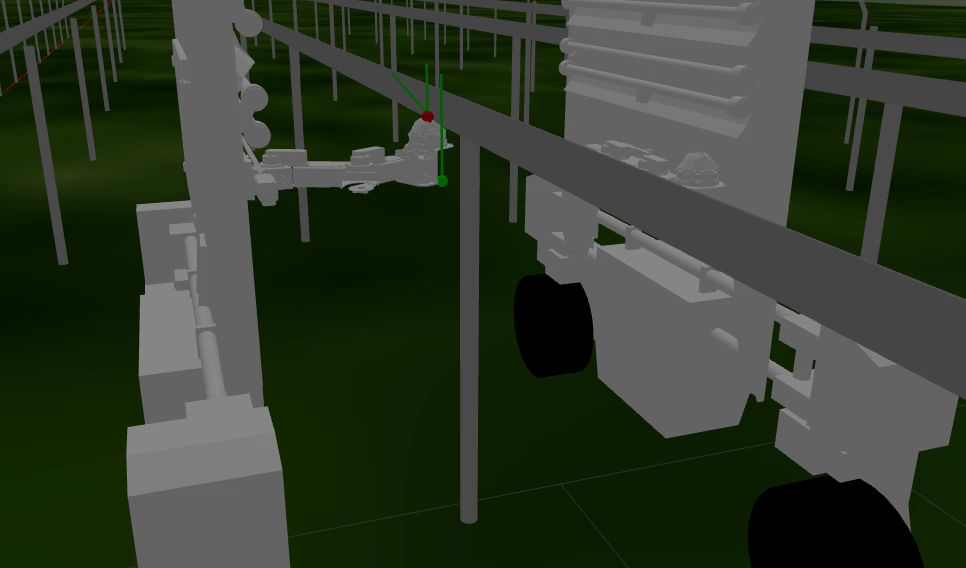}
         \label{fig:scara_promp_sim1g}
     \end{subfigure}\hspace{10pt}
     \begin{subfigure}[t]{0.4\columnwidth}
     \centering
         \includegraphics[trim=270 220 480 190, clip, width=\textwidth]{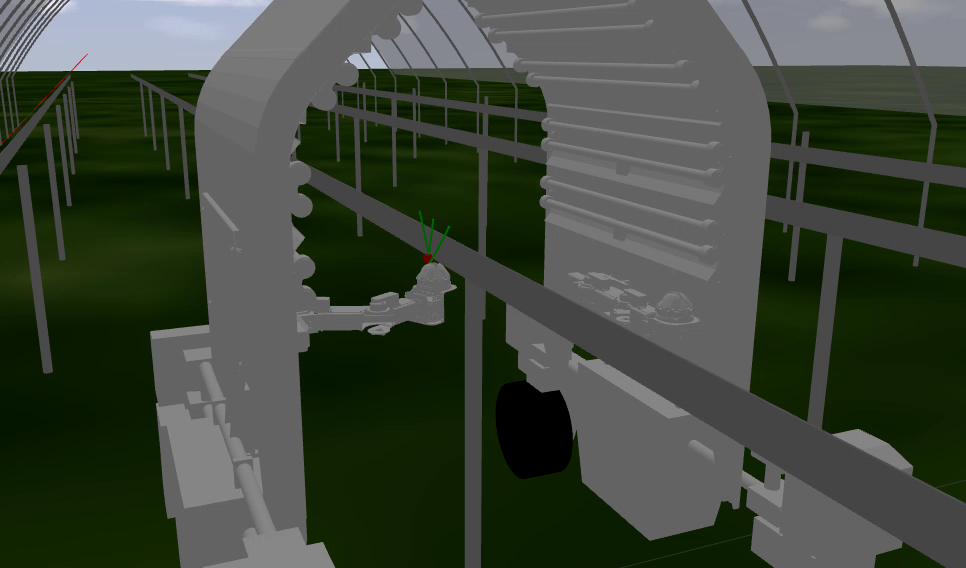}
         \label{fig:scara_promp_sim2g}
     \end{subfigure}
     \vspace{-15pt}
\caption{ProMP generation leveraging the basic framework in \cite{paraschos2013probabilistic}. (right) conditioning the $ProMP$ on the initial pose and final desired pose, (left) conditioning the $ProMP$ on the initial pose, camera positioning below the target and outside the cluster radius, and the desired target fruit pose.}\label{fig:promp_sim}
\end{figure}

\noindent We consider in a first place a case where ripe fruits are located very close to a static obstacle, the table-top. This test allows to showcase the performance of a basis $ProMP$ planner and the extent to which an advanced planner is needed to approach complex real scenarios. In fig.~\ref{fig:promp_sim} we report time frames to showcase the performance of running ProMP on a Scara arm of Thorvald. The complete simulations for this case scenario and the following ones are reported in the media materials\footnote{https://youtu.be/klAaQ6Ocwb8}. Fig.~\ref{fig:promp_sim} (right) reports the performance of a $ProMP$ trajectory generation while conditioning the learnt primitive on the initial task space pose and the desired target pose. It can be clearly seen the collision between the gripper and the table-top before any swallowing and stem cut could have been made. In addition, the left hand stem of the neighbour unripe fruit is overtaken by the gripper causing a change in the target fruit pose because of stems collision. Fig.~\ref{fig:promp_sim} (left) reports the performance of a $ProMP$ generation while conditioning the learnt primitive on an additional point below the target and $10cm$ below the cluster radius. It illustrates again a collision with the table-top while the gripper overtakes the right hand stem causing a change in the target fruit pose.\\

\noindent \paragraph*{\textit{\textbf{B. Scenario-II}}: \textbf{I-ProMP for picking ripe fruits in cluster with Scara arm}}

\begin{figure}
     \begin{subfigure}[t]{0.355\columnwidth}
     \centering
         \fbox{\includegraphics[trim=900 400 550 280, clip, width=\textwidth]{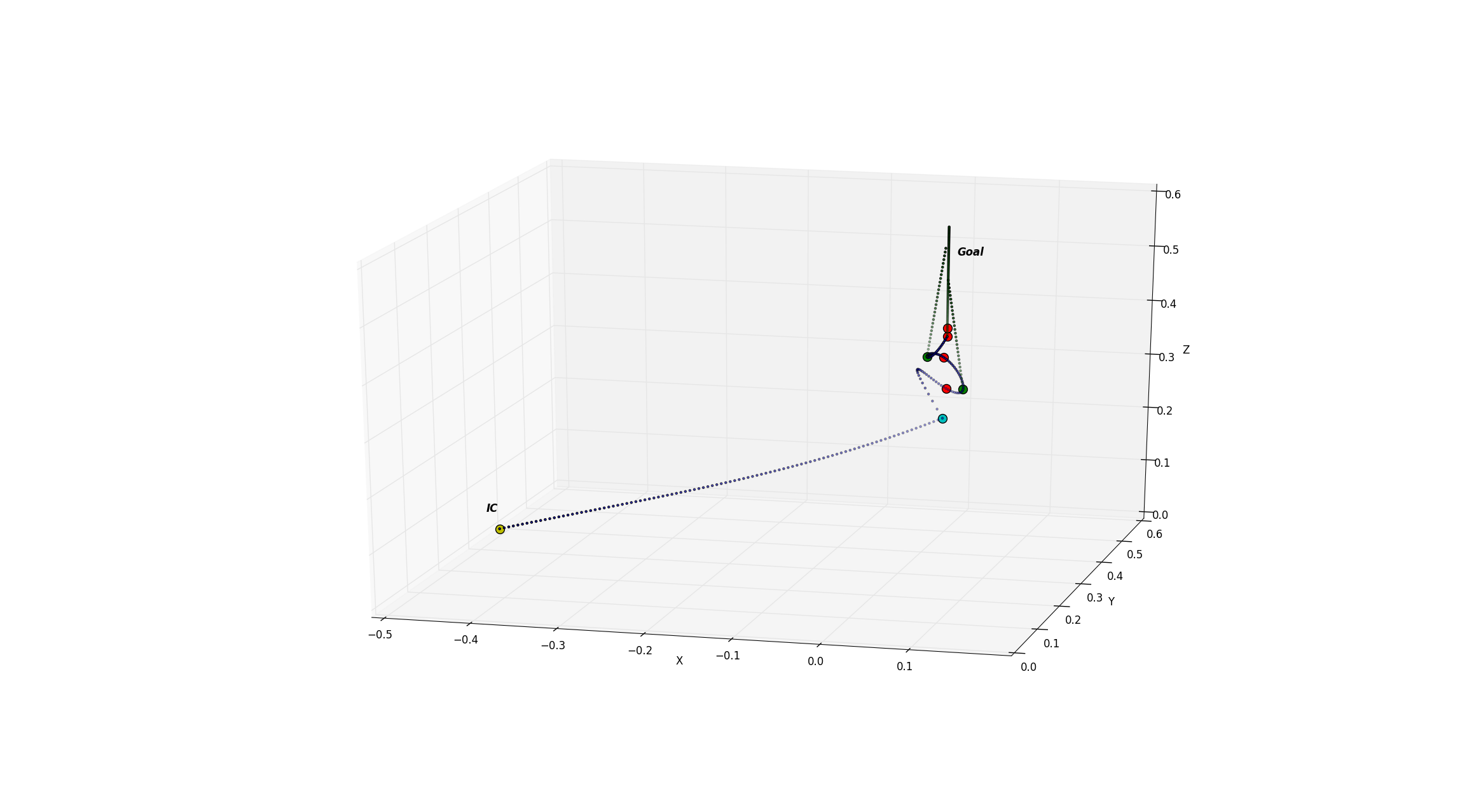}}
         \label{fig:intpromp_py}
     \end{subfigure}\hspace{10pt}
     \begin{subfigure}[t]{0.55\columnwidth}
     \centering
         \includegraphics[trim=225 230 500 165, clip, width=\textwidth]{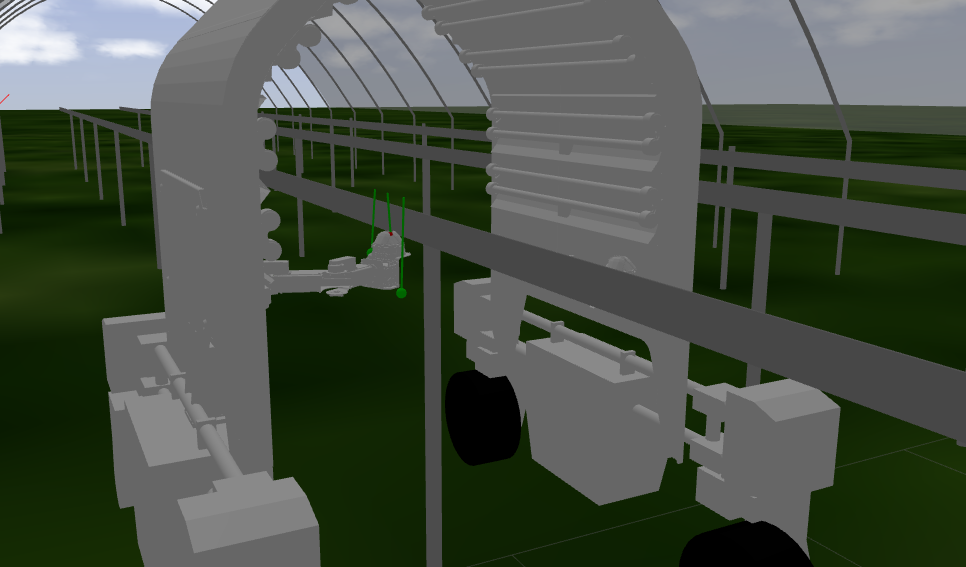}
         \label{fig:intpromp_sim}
     \end{subfigure}
     \vspace{-10pt}
\caption{(right) I-ProMP generation as per \cite{mghamesinteractive}, leading to pushing actions on unripe fruits (illustrated in red spheres with no stems for clarity, while green spheres represent the updated pose of the unripe fruits resulting from a systematic pushing direction, and the ripe fruit is the red sphere with stem) sampled from a Gaussian distribution.
(left) time frame from the simulated I-ProMP showing collision of a fingers-like gripper with a table-top meanwhile attempting to pick a target fruit nearby the table. }\label{fig:intpromp}
\end{figure}

\noindent In a following step, we test the I-ProMP as formulated in~\cite{mghamesinteractive} on a cluster with unripe fruits occluding the ripe target from below (case of 2 unripe fruits) while shifting the target pose to near the table-top as opposed to the configuration reported in~\cite{mghamesinteractive}. In fig.~\ref{fig:intpromp} (right) we report the I-ProMP generation for the aforementioned cluster configuration, and in fig.~\ref{fig:intpromp} (left) we report a time frame of the simulation conducted in a digital-twin poly-tunnel built in Gazebo. The latter figure illustrates the collision of the gripper with the table-top before being able to swallow the target and after having pushed away the occluding elements. The reader can refer to the media attachment for the visualisation of the complete simulation.\\

\noindent \paragraph*{\textit{\textbf{C. Scenario-III}}: \textbf{OptI-ProMP for picking ripe fruits in cluster with Franka Emika arm - Case study with right and left hand cluster}}

\begin{figure*}

     \begin{subfigure}[t]{0.25\textwidth}
     \centering
         \includegraphics[trim=5 0 50 0, clip, width=\textwidth]{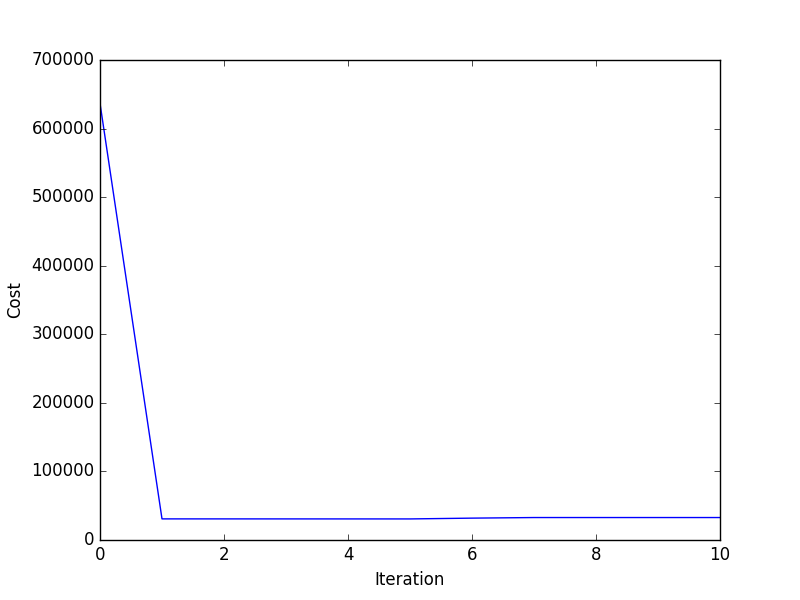}
         \caption{}
         \label{fig:optpromp_clus_push}
     \end{subfigure}\hspace{0.1pt}
     \begin{subfigure}[t]{0.24\textwidth}
     \centering
         \includegraphics[trim=15 0 50 0, clip, width=\textwidth]{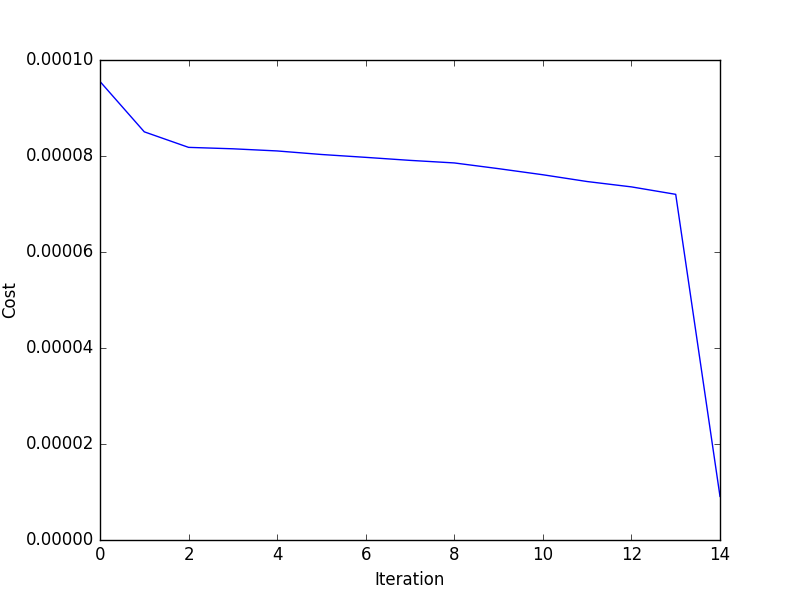}
         \caption{}
         \label{fig:optpromp_clus_obs}
     \end{subfigure} \hspace{0.1pt}
    \begin{subfigure}[t]{0.24\textwidth}
     \centering
         \includegraphics[trim=10 0 50 0, clip, width=\textwidth]{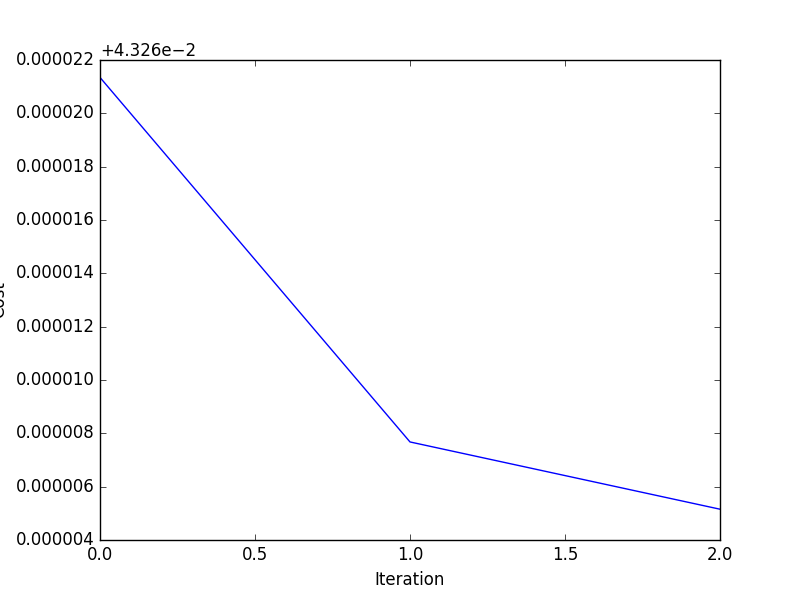}
         \caption{}
         \label{fig:optpromp_table_obs}
     \end{subfigure}\hspace{0.1pt}
    \begin{subfigure}[t]{0.235\textwidth}
     \centering
         \includegraphics[trim=25 0 50 0, clip, width=\textwidth]{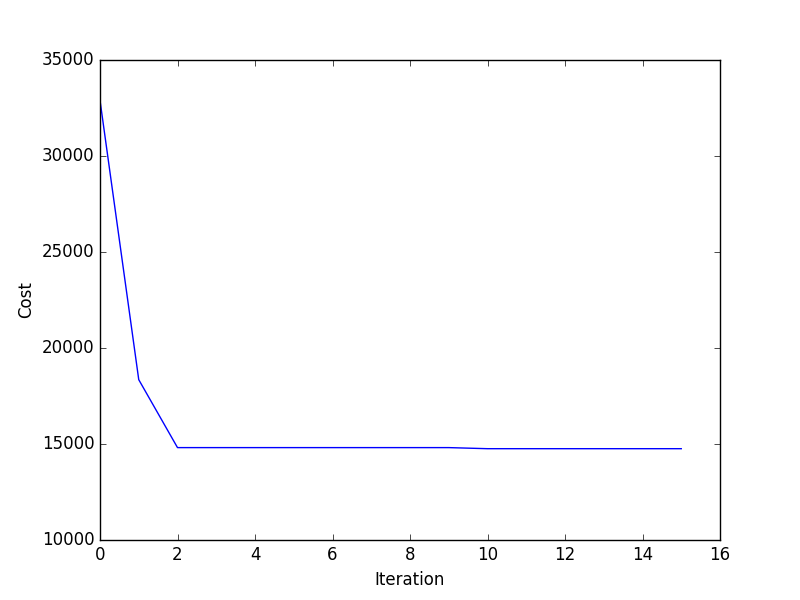}
         \caption{}
         \label{fig:optpromp_table_push}
     \end{subfigure}
     \newline

          \begin{subfigure}[b]{0.222\textwidth}
     \centering
         \fbox{\includegraphics[trim=445 415 210 70, clip, width=\textwidth]{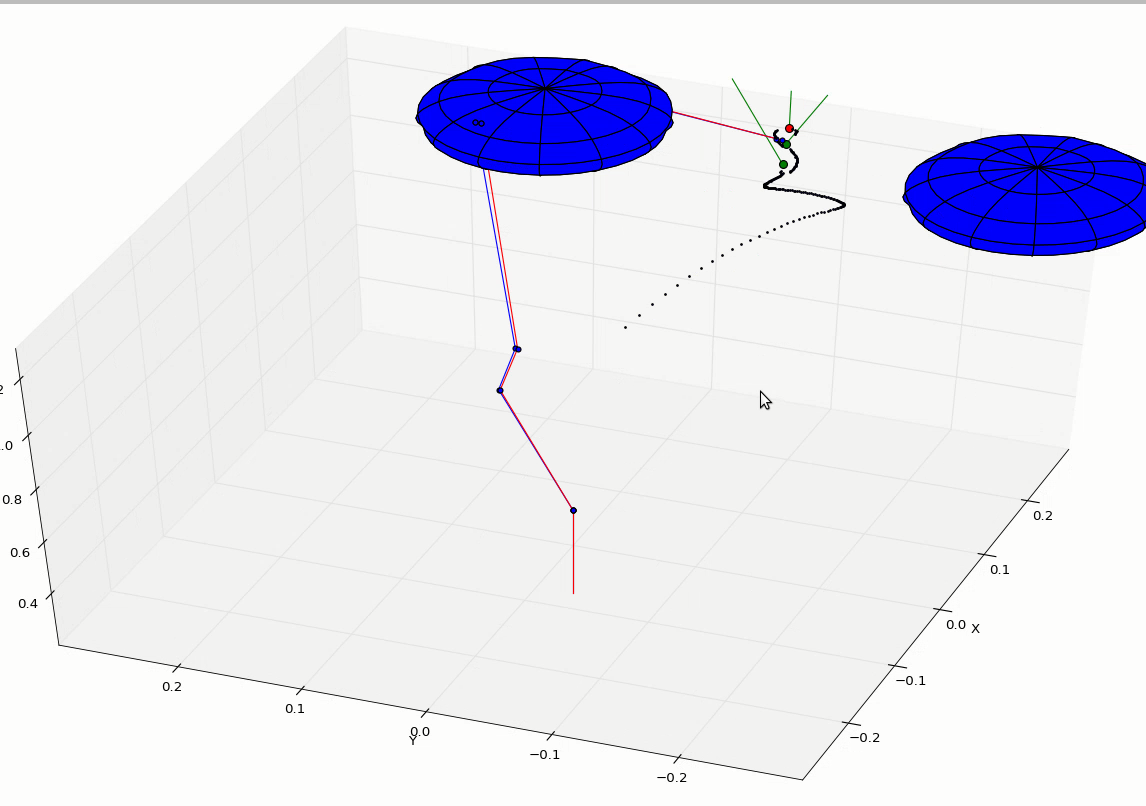}}
         \caption{}
         \label{fig:sim_optpromp_clus_push}
     \end{subfigure}\hspace{15pt}
     \begin{subfigure}[b]{0.223\textwidth}
     \centering
         \fbox{\includegraphics[trim=170 300 710 290, clip, width=\textwidth]{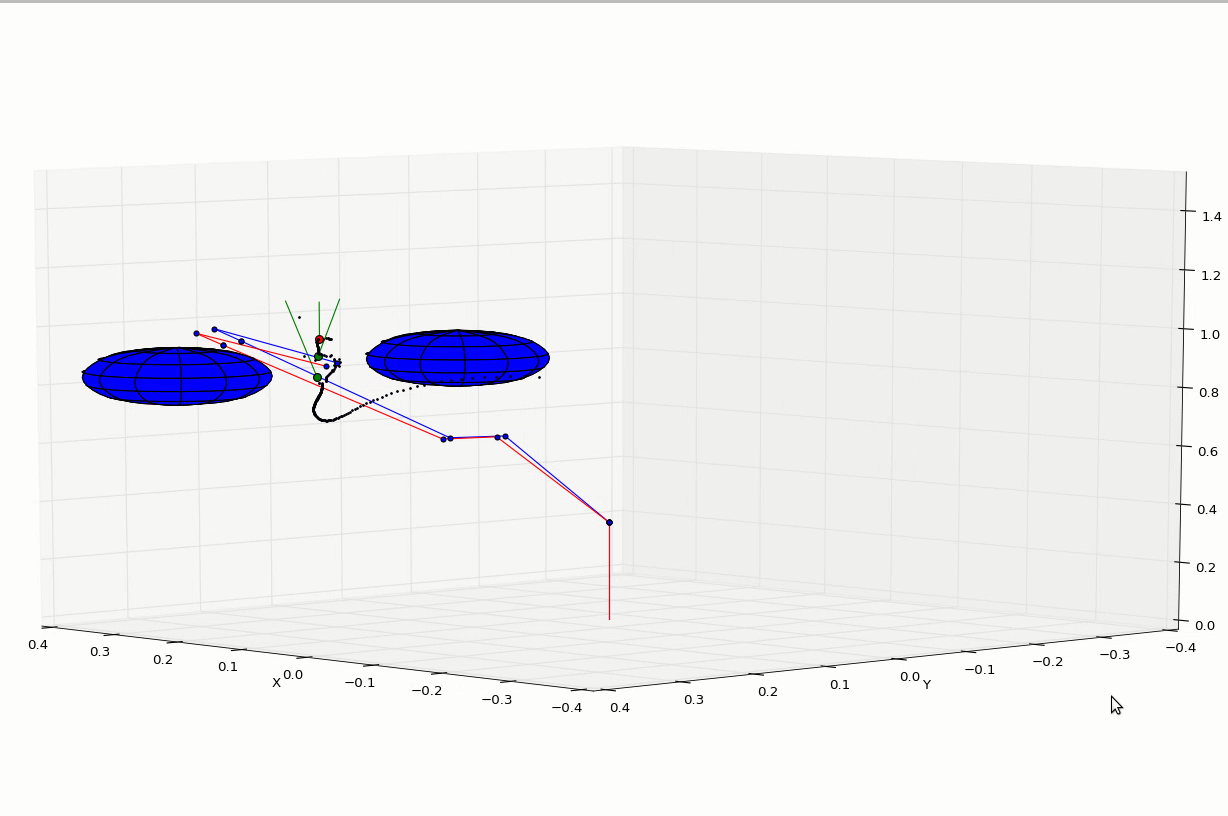}}
         \caption{}
         \label{fig:sim_optpromp_clus_obs}
     \end{subfigure} \hspace{10pt}
    \begin{subfigure}[b]{0.222\textwidth}
     \centering
         \fbox{\includegraphics[trim=250 350 700 240, clip, width=\textwidth]{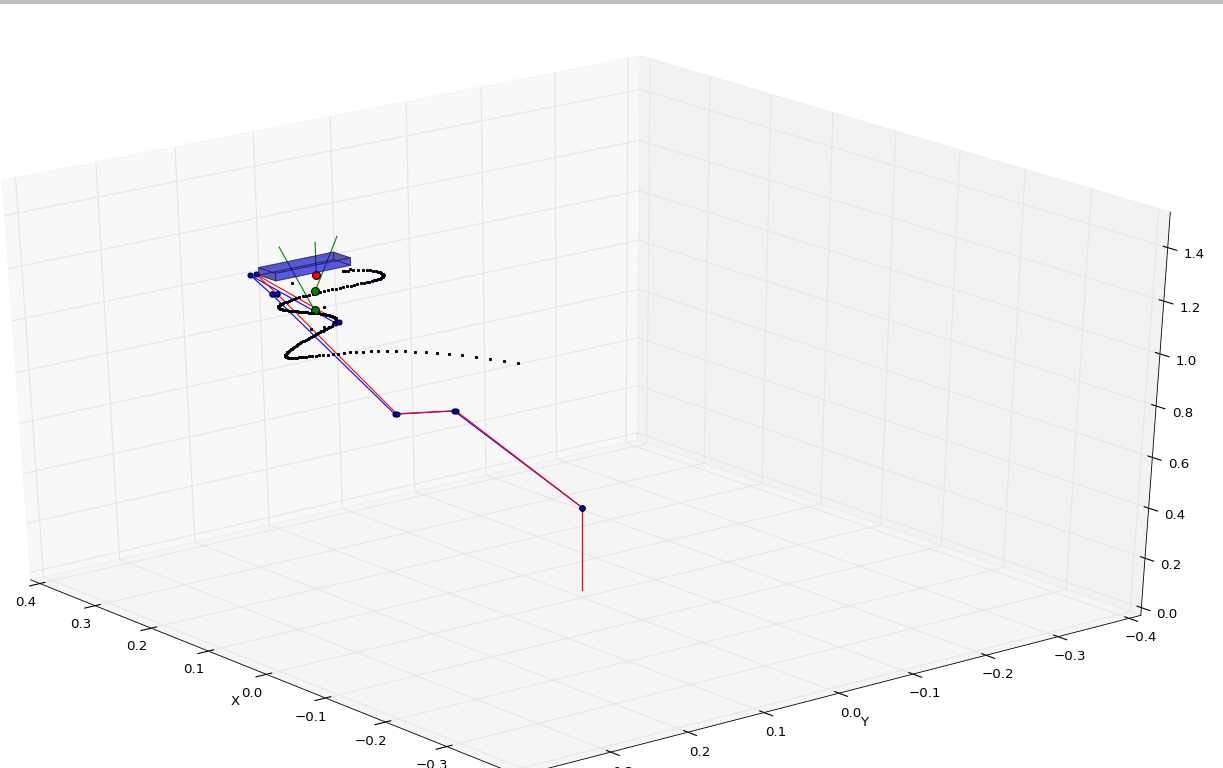}}
         \caption{}
         \label{fig:sim_optpromp_table_obs}
     \end{subfigure}\hspace{10pt}
    \begin{subfigure}[b]{0.22\textwidth}
     \centering
         \fbox{\includegraphics[trim=520 370 570 240, clip, width=\textwidth]{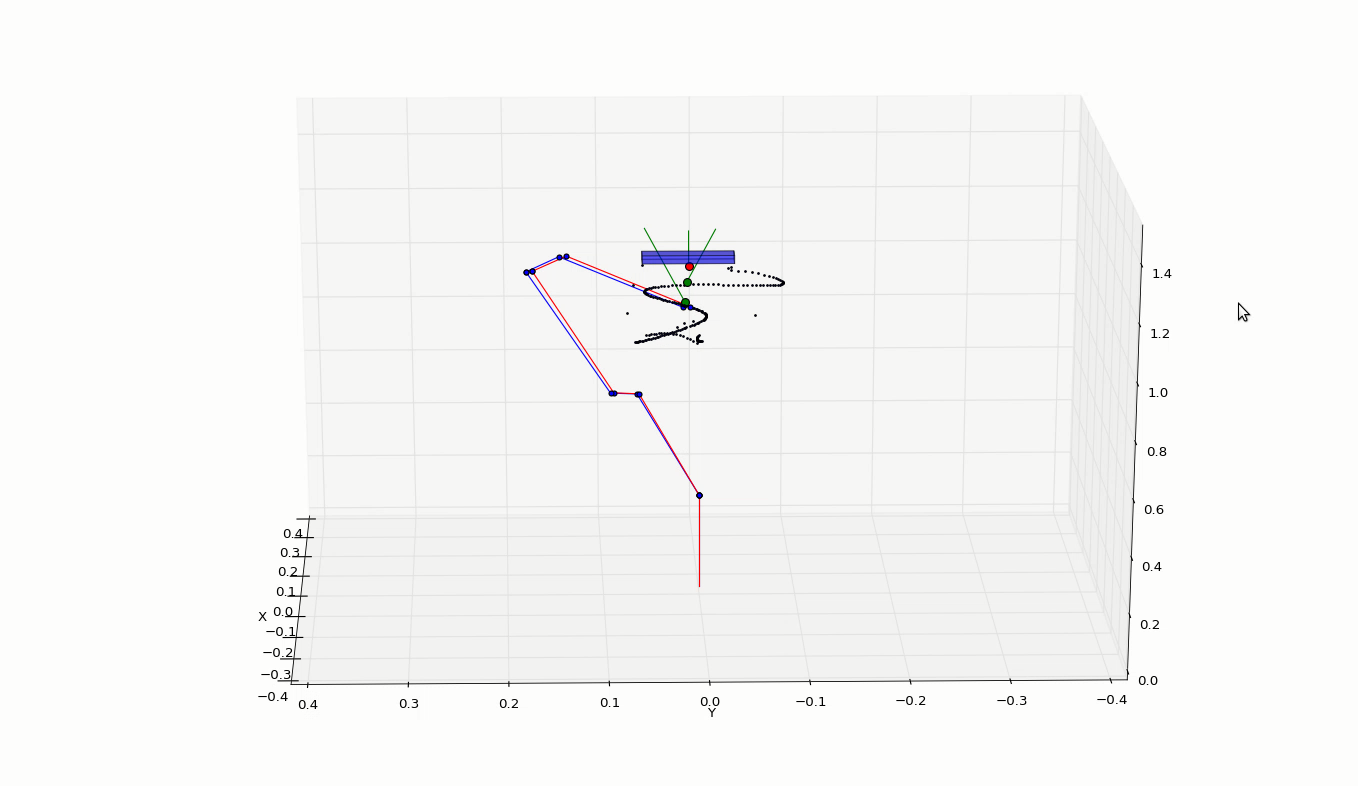}}
         \caption{}
         \label{fig:sim_optpromp_table_push}
     \end{subfigure}

\caption{OptI-ProMP generation: (a)-(d) optimization cost over iterations, (e)-(h) simulated initial (red) and final optimised (blue) joints trajectory, corresponding to (a)-(d). The ripe fruit is illustrated in red while unripe ones in green, each with its corresponding stem inclination selected for a configuration case study. (a)-(e) minimizes a pushing cost, (b)-(f) minimizes sequentially a collision cost, with right and left clusters being the obstacles, (c)-(g) minimizes a collision cost, with table-top being the obstacle, while (d)-(h) minimizes sequentially a pushing cost for the case scenario of (c)-(g).}\label{fig:optpromp}
\end{figure*}

\noindent Since the I-ProMP framework was not successful in handling environmental constraints, we test in the following the proposed OptI-ProMP optimisation framework on a higher-dexterity level manipulator (Franka Emika with 7-dofs) while binding ourselves in this work to the consideration of the omni-directional Franka gripper. Under the latter approach and as a first test, we consider the static obstacles to be the right and left hand clusters. We solve the optimisation problem in a sequential setting, using the trust-region optimiser to minimise the pushing cost and the L-BFGS-B optimiser to minimise the static obstacles cost (or what is considered static) and the velocity cost. Both optimisers converge to local minima. Figures~\ref{fig:optpromp_clus_push} and  \ref{fig:sim_optpromp_clus_push} report respectively the cost after 10 iterations and a time frame from the simulated initial (red manipulator links) and optimised (blue manipulator links following the blue dotted trajectory) trajectories in which we can see a small shift of the ee from the center of the unripe fruit located right underneath the target. The pushing cost converges from $\sim620,000$ to a value below $40,000$. We note that the shift in fig.~\ref{fig:sim_optpromp_clus_push} is small and hence the reader can refer to the attached media for a better visualisation. The optimised trajectory output from the trust-region optimiser is now used to initialise the L-BFGS-B optimiser for collision cost minimisation. Figures~\ref{fig:optpromp_clus_obs} and \ref{fig:sim_optpromp_clus_obs} report respectively the collision cost after 15 iterations and a time frame from the simulated initial and optimised trajectories in which we can see a drift of the blue optimised configuration from the initialised red robot configuration. The pushing cost converges from $\sim9.5e-05$ to a value below $1.0e-05$.
We note that the performance simulation is reported without pose update of the pushable objects, leaving the physics-based Gazebo testing for future works.\\

\begin{table}[t]
\centering
\begin{tabular}{l|l|l|l}
 & $M_c$ & $M_p(X_{n^1})$ & $M_p(X_{n^2})$  \\
 \hline
R/L clusters & 0.157 & -0.01 & 0.001968  \\
\hline
Table-top & 0.0352 & -0.0084 & 0.03379   \\
\end{tabular} 
\caption{OptI-ProMP performance measure.}
\label{tab:measure}
\end{table}

\noindent \paragraph*{\textit{\textbf{D. Scenario-IV}}: \textbf{OptI-ProMP for picking ripe fruits in cluster with Franka Emika arm - Case study with adjacent table-top}}

\noindent Under the OptI-ProMP approach, we consider in a second test static obstacles to be table-top. The sequential optimisation is reverted in this case, in the sense that we minimise the collision cost before the pushing cost. The rational goes back to the importance of minimizing last the cost with higher weight given the situation in hand. Hence, we consider the closeness of the table-top a hard constraint with higher importance when compared to the nearby clusters closeness. Figures~\ref{fig:optpromp_table_obs} and  \ref{fig:sim_optpromp_table_obs} report respectively the cost after 3 iterations and a time frame from the simulated initial and optimised trajectories in which we can see a drift of the blue optimised configuration from the initialised red robot configuration. The collision cost converges from $\sim2.1e-05$ to a value $\sim5.0e-06$. The output optimal trajectory from the L-BFGS-B optimiser is now used to initialise the trust-region optimiser for pushing cost minimisation. Figures~\ref{fig:optpromp_table_push} and \ref{fig:sim_optpromp_table_push} report respectively the cost after 15 iterations and a time frame from the simulated initial and optimised trajectories in which we can see a drift of the ee from the center of the lowest unripe fruit. We note that the optimised trajectory in this specific case may not outperform the initialised one when deployed on real system and that's because the initialised trajectory shows sub-optimal drifting behavior from the unripe fruit center. However, we can see that the optimised trajectory drifts from the initial one in the same direction, avoiding to flip the pushing direction at the center level. The pushing cost converges from $\sim33,000$ to a value below $15,000$.\\

\noindent \textbf{Performance Measure}
A quantitative assessment of OptI-ProMP performance is reported in table~\ref{tab:measure} for all the above cases. The collision cost measure is formulated as $M_c = min \hspace{5pt} d(u_{s,d}, \{\mathcal{O}\})$ where $\{\mathcal{O}\}$ is the set of static obstacles, whereas the push cost measure is formulated as $M_p = proj_y \, \mathbf{d}(X_{ee},X_n)$ where $X_{ee,n}$ is the pose of the ee and neighbour object, respectively. We verify hereafter that $M_c > 0.1$ for right (R) and left (L) hand clusters case with clusters radius $10cm$, $M_c > 0.03$ for table-top case whose depth is $3cm$, and $M_p$ is of opposite sign to the projection of the stem vector ($\mathbf{S}$) on $y=\{0, 1, 0\}$ originating from $X_n$. Two neighbour unripe fruits ($n^1$ and $n^2$, farthest to nearest to target respectively) are present with projections $proj_y \, \mathbf{S}_{n^1} > 0$ and $proj_y \, \mathbf{S}_{n^2} < 0$ on $y$ as can be seen from fig.~\ref{fig:intro} with $y$ along $y_g$.



%% file: conclusion.tex
\section{Conclusion}
In this work, we propose a new multi-objective optimization framework for handling object reaching in 3-d cluttered environments (and possibly 2-d). The framework leverages the local planner CHOMP with a $ProMP$ initialisation to extend the objective and constraints functions to pushable dynamic objects while considering connections between target and neighborhood. The framework was implemented on case scenarios from an agricultural environment where high-functionalities gripper designs may fail to complete the task. Results from simulation with Scara and Franka Emika arms show success of high-dexterity robot body in reaching for an object in clutter by generating pushing actions systematically on occluding elements meanwhile avoiding static obstacles in its surrounding.
Future works will be devoted to (a) test the framework on the real robot, (b) finalising the stem detection and inclination estimate module which we started to work on leveraging abstract shapes dataset generated on the fly and from simulations to train a mask-RCNN detection framework,
and (c) comparing the OptI-ProMP performance to a deep learning approach.
